\documentclass{bmvc2k}
\usepackage[T1]{fontenc}
\usepackage{multirow}
\usepackage{makecell}
\usepackage{booktabs}
\usepackage{graphicx}
\usepackage{amsmath}
\usepackage{adjustbox}
\usepackage{algorithm}
\usepackage{algpseudocode}
\usepackage{arydshln}
\usepackage{xcolor}
\usepackage{amssymb}
\usepackage{commath}

\usepackage{graphicx}
\usepackage{xcolor}

\title{3D Point Cloud Network Pruning: When Some Weights Do not Matter}

\newcommand\blfootnote[1]{%
  \begingroup
  \renewcommand\thefootnote{}\footnote{#1}%
  \addtocounter{footnote}{-1}%
  \endgroup
}

\addauthor{Amrijit Biswas$^\ast$}{amrijit.biswas01@northsouth.edu}{1}
\addauthor{Md. Ismail Hossain$^\ast$}{ismail.hossain2018@northsouth.edu}{1}
\addauthor{M M Lutfe Elahi}{lutfe.elahi@northsouth.edu}{1}
\addauthor{Ali Cheraghian}{ali.cheraghian@data61.csiro.au}{2,3}
\addauthor{Fuad Rahman}{fuad@apurbatech.com}{4}
\addauthor{Nabeel Mohammed}{nabeel.mohammed@northsouth.edu}{1}
\addauthor{Shafin Rahman}{shafin.rahman@northsouth.edu}{1$^{\dagger}$}

\addinstitution{
Department of Electrical and\\Computer Engineering,\\North South University, Bangladesh
}
\addinstitution{
 Data61-CSIRO, Australia
}

\addinstitution{
The Australian National University,\\
 Canberra, Australia
}

\addinstitution{
Apurba Technologies, Sunnyvale, \\ CA 94085, USA
}

\runninghead{Biswas et al.}{3D Point Cloud Network Pruning}

\begin{document}

\maketitle
\blfootnote{$^\ast$ Equal Contribution.}
\blfootnote{$^{\dagger}$ Corresponding author.}

\begin{abstract}

A point cloud is a crucial geometric data structure utilized in numerous applications. The adoption of deep neural networks referred to as Point Cloud Neural Networks (PCNNs), for processing 3D point clouds, has significantly advanced fields that rely on 3D geometric data to enhance the efficiency of tasks. Expanding the size of both neural network models and 3D point clouds introduces significant challenges in minimizing computational and memory requirements. This is essential for meeting the demanding requirements of real-world applications, which prioritize minimal energy consumption and low latency. Therefore, investigating redundancy in PCNNs is crucial yet challenging due to their sensitivity to parameters. Additionally, traditional pruning methods face difficulties as these networks rely heavily on weights and points. Nonetheless, our research reveals a promising phenomenon that could refine standard PCNN pruning techniques. Our findings suggest that preserving only the top p\% of the highest magnitude weights is crucial for accuracy preservation. For example, pruning ~99\% of the weights from the PointNet model still results in accuracy close to the base level. Specifically, in the ModelNet40 dataset, where the base accuracy with the PointNet model was 87. 5\%, preserving only 1\% of the weights still achieves an accuracy of 86.8\%. Codes are available in: \url{https://github.com/apurba-nsu-rnd-lab/PCNN_Pruning}
\end{abstract}


\section{Introduction}



The ability to analyze and comprehend 3D data is becoming increasingly vital in various industries, such as autonomous driving \cite{li2023mseg3d}, robotics \cite{liu2023path}, augmented and virtual reality \cite{urlings2023views}, and computational biology \cite{pajaziti2023shape}. The widespread availability and diverse application range of 3D data have contributed to its growing importance. With the continued advancement of deep learning technologies, researchers are exploring innovative ways to process and interpret 3D data effectively. This marks the dawn of a new era in 3D deep learning research, commonly referred to as Point Cloud Neural Networks (PCNNs) \cite{goyal2021revisiting}. Early innovations, such as PointNet \cite{qi2017pointnet}, have laid the groundwork for developing deep learning architectures specifically designed to process raw point clouds. Subsequent developments, including PointCNN \cite{li2018pointcnn}, PointConv \cite{wu2019pointconv}, PointCLIP \cite{zhang2022pointclip}, and Uni3d \cite{zhou2023uni3d}, have introduced improved strategies for capturing minute geometric details and enhancing matching accuracy through advanced convolution and attention mechanisms. 
The growing need for innovative deep-learning solutions demands efficient models and reasoning with 3D data. However, as models become more complex over time, the number of parameters increases significantly. This substantial increase can lead to heightened latency and computational constraints, posing significant challenges to the efficient deployment and operation of models.

\begin{figure}[t!]
\centering
\begin{center}
\includegraphics[width=1\linewidth]{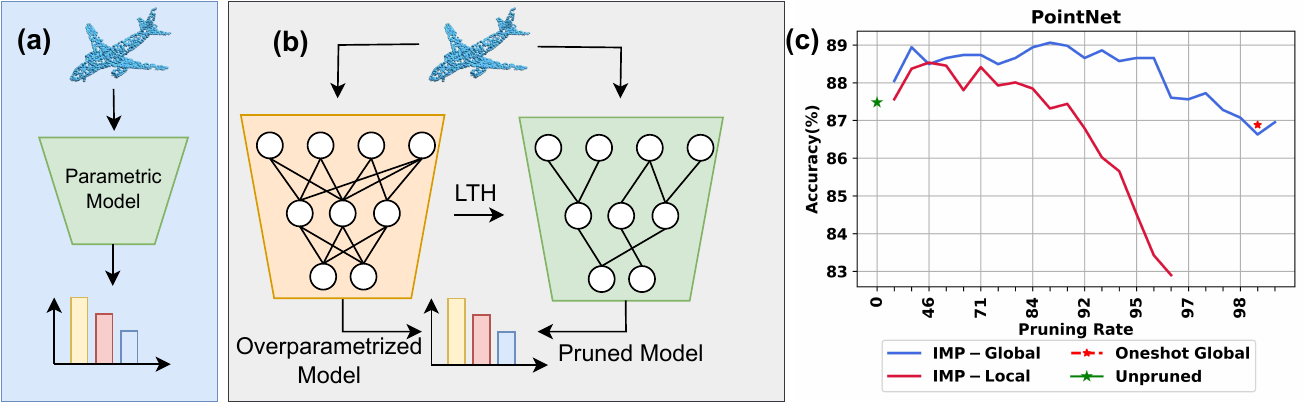}
\end{center}
\vspace{-2em}
\caption{\small This paper deals with pruning 3D  Point Cloud Neural Networks (PCNNs) for faster inference, especially on edge devices. \textbf{(a)} In 3D point cloud literature, traditionally, researchers carefully design less parametric deep models for this purpose (e.g., Spherical CNNs \cite{esteves2018learning}, Dense Point \cite{liu2019densepoint}, KCNet \cite{shen2018mining}, Point PN \cite{zhang2023parameter}). \textbf{(b)} In contrast, we iteratively prune popular PCNNs (PointCNN \cite{li2018pointcnn}, DGCNN \cite{wang2019dynamic}, PointConv \cite{wu2019pointconv}) to produce task-specific subnetworks outperforming over-parameterized models. \textbf{(c)} Our results demonstrate the capability of the LTH  to extract highly sparse subnetworks, termed "winning tickets," from an over-parameterized model. These winning tickets can achieve up to 99\% sparsity, retaining only 1\% of the original weights while still attaining accuracy levels comparable to the base model.
}
\label{fig:intro}
\end{figure}

To address the challenge, four standard methods are commonly 
utilized: weight quantization \cite{yang2019quantization}, sparsity through regularization \cite{girosi1995regularization}, knowledge distillation \cite{zhou2021distilling}, and network pruning \cite{blalock2020state}. 
Each method has its own advantages and disadvantages. While they aim to improve efficiency and reduce computational demands, each technique may also potentially compromise accuracy or introduce other trade-offs that need to be carefully considered. In that context, network pruning is a technique, especially unstructured pruning, that has yet to be extensively explored within this specific 3D domain (see Figure \ref{fig:intro}). Unstructured pruning \cite{liao2023can} has emerged as a promising technique for model compression in other domains. It involves identifying and eliminating redundant or irrelevant weights from the network, thereby creating a more compact model without sacrificing accuracy. Although this method, especially the recently proposed Lottery Ticket Hypothesis (LTH) \cite{frankle2018lottery}, has found widespread application in various fields, its effectiveness in compressing 3D models has not yet been tested.

LTH suggests that there are trainable subnetworks within larger neural networks, termed "winning tickets," that can achieve or exceed the original performance. These subnetworks, identifiable through iterative pruning, exhibit potential for task transferability \cite{chen2021elastic}, sparsity, enhanced performance, and convergence. Specifically, our aim is to address the following research questions:
\textbf{(a)} Can we find winning ticket subnetworks within overparametrized 3D shape classification networks that maintain or exceed the original network's performance while being significantly sparser? 
\textbf{(b)} What are the optimal pruning strategies and techniques to effectively identify winning tickets in 3D models, considering the unique challenges posed by high-dimensional and geometrically complex data? 
\textbf{(c)} Can the winning ticket subnetworks obtained from a 3D shape classification task or dataset be transferred to different but related tasks while preserving high accuracy, thus demonstrating the transferability of these subnetworks? We found winning tickets within overparameterized PCNNs that maintained or sometimes exceeded the original performance while being significantly sparser. IMP \cite{zhang2022advancing}
and global one-shot pruning \cite{chen2021only} emerged as optimal pruning strategies for effectively handling the high-dimensional and geometrically complex nature of 3D data. In particular, we found that these winning ticket subnetworks exhibited transferability across different but related 3D shape classification tasks, preserving high accuracy. These investigations have significant potential to propel the field of 3D model compression. 
To validate our hypotheses, we employ versatile 3D point cloud architectures like PointNet \cite{qi2017pointnet}, DGCNN \cite{wang2019dynamic}, and PointCNN \cite{li2018pointcnn}. Extensive experiments are conducted on challenging 3D datasets: ModelNet40 \cite{wu20153d}, ScanObjectNN \cite{uy-scanobjectnn-iccv19}, and ShapeNetCore \cite{chang2015shapenet}, encompassing diverse 3D objects and scenarios. 
The contributions of this paper are:

\begin{itemize}

\item  Development of efficient sparse, task-specific subnetworks outperforming over parameterized models (e.g., PointCNN \cite{li2018pointcnn}, DGCNN~\cite{wang2019dynamic}, PointConv \cite{wu2019pointconv}) for efficient 3D deployment. 

\item Comprehensive analysis of sparse subnetwork characteristics and performance of most popular models across diverse datasets (ModelNet40 \cite{wu20153d}, ScanObjectNN \cite{uy-scanobjectnn-iccv19}, and ShapeNetCore \cite{chang2015shapenet}).

\item We establish that one-shot global pruning at a highly high 99\% sparsity level can attain comparable accuracy to over-parameterized 3D models, substantially reducing parameter count and computational requirements. The rest of the 1\% critical weights play a vital role in performance.

\end{itemize}

\section{Related Works}

\noindent\textbf{3D Point Cloud Neural Networks:} Existing methods for classifying 3D shapes can be broadly categorized into multi-view, volumetric, and point-based methods. Multi-view-based models, such as the Multi-View Convolutional Neural Network (MVCNN) \cite{su2015multi}, convert unstructured 3D point clouds into 2D images from different perspectives. Extracted features from images are then combined to achieve a complete global representation. Volumetric-based techniques (VoxelNet \cite{zhou2018voxelnet}, OctNet \cite{riegler2017octnet}, and Octree-based CNN \cite{wang2017cnn}) convert point clouds into a structured format, employing representations such as voxels or octrees. Subsequently, these methods utilize 3D Convolutional Neural Network (CNN) models, including but not limited to performing shape classification. 
In contrast, point-based classification methods directly handle the raw, unstructured point clouds. These can be divided into several subcategories: pointwise Multilayer Perceptron (MLP), convolutional-based, graph-based, and hierarchical data structure-based methods. For instance, PointNet \cite{qi2017pointnet} exemplifies the pointwise MLP approach by independently extracting features from each point through multiple MLP layers and aggregating them via a max pooling layer to capture global shape features. Meanwhile, PointCNN \cite{li2018pointcnn}, 
demonstrate the application of convolutional approaches tailored to point cloud data. Graph-based methods (DGCNN \cite{wang2019dynamic}) treat each point in the cloud as a vertex in a directed graph, leveraging the relationships between points to facilitate shape classification. This paper investigated 3D classification models from each subcategory.

\noindent\textbf{3D model Compression:} Recent advances in 3D PCNN compression research have predominantly focused on two main areas: geometry compression \cite{wang2021lossy} and attribute compression \cite{fang20223dac}. The literature primarily features methods utilizing convolution-based autoencoders (CNN-based AE), fully connected neural networks (FCNN), and multilayer perceptrons (MLP). Among the various strategies explored, channel pruning \cite{huang2023cp3} emerges as a significant approach, drawing inspiration from the principles of 2D channel pruning \cite{hua2019channel}. Application of this method to the ModelNet40 \cite{wu20153d} has demonstrated that a pruning rate of 58\% can maintain, or in some cases improve, accuracy relative to the benchmark PCNN \cite{ vinodkumar2023survey, cheraghian2019mitigating}. Moreover, Point Distribution-Aware Pruning \cite{lee2023not} strategically prunes less significant neighborhood voxels to enhance compression efficiency. Some other works \cite{zhang2023parameter, nguyen2023lossless} investigated the potential of employing PCNNs with a minimal number of parameters for 3D shape classification. Proposals for non-parametric networks \cite{zhang2023parameter} aimed at learning 3D shapes have been introduced. Despite these efforts, such non-parametric models have shown limited success on the ScanObjectNN 3D dataset \cite{uy-scanobjectnn-iccv19}. However, with a reduced parameter count of 0.8M, these models have achieved notable accuracy on the ScanObjectNN dataset, highlighting the trade-offs and potential pathways for further optimization in PCNN compression techniques. Yet, the utilization of LTH on 3D model compression still needs to be explored.

\noindent\textbf{Lottery Ticket Hypothesis:} The LTH \cite{frankle2018lottery} has unleashed a new dimension in the domain of neural network pruning. Later, \cite{frankle2019stabilizing} introduced the term "late resetting," which facilitates the LTH suitable for deep models. However, \cite{zhou2019deconstructing} endorsed that weight resetting can be ambiguous. The reason behind the working process of experimental LTH was explored by \cite{frankle2020linear}. Further, the theoretical proof of LTH was also evaluated by \cite{malach2020proving}. Authors \cite{sapkota2024distributionally} showed that a winning ticket distribution can be present in a neural network. The long-term training issue of the LTH was addressed in \cite{jaiswal2023instant}. Furthermore, the concept of "transfer ticket hypothesis" \cite{iofinova2022well} allows training a network by using a sparse network (winning ticket) generated from another dataset. Authors \cite{hansen2021lottery} demonstrated that the transferability of the winning tickets does not cause overfitting issues. LTH technique is successfully applied in several computer vision applications, e.g., supervised pre-training \cite{ he2023transferring}, object recognition \cite{girish2021lottery}, vision-language models \cite{gan2022playing}, which inspired us to investigate LTH on 3D point cloud shape classification.

\section{Lottery Ticket Hypothesis for 3D Shape Classification}

\noindent\textbf{Preliminaries:}
The LTH suggests that dense, randomly initialized neural networks contain subnetworks (known as "winning tickets") that can match the performance of the original network when trained in isolation. In the context of 3D shape classification, we aim to prove the existence of winning tickets within a neural network $\mathcal{F}(\mathbf{x}; \theta)$ that maps 3D shapes $\mathbf{x} \in \mathcal{X}$ to class labels $\mathcal{Y}$ using parameters $\theta$ initialized from a distribution $\mathcal{D}_\theta$.
To find the winning ticket subnetwork, we employ an iterative pruning and retraining process:
\begin{itemize}\setlength{\itemsep}{-0.2em}

    \item[1.] Initialize a binary mask $\mathbf{m} \in \{ {0, 1}\} ^{|\theta|}$ with all ones: $\mathbf{m} = \mathbf{1}$.

    \item[2.] Train the network $\mathcal{F}(\mathbf{x}; \mathbf{m} \odot \theta)$ for one cycle. 
    \item[3.] Prune a percentage $p$ of the weights by setting the corresponding entries in $\mathbf{m}$ to zero based on a pruning criterion. Where ${m}_i = 0$, if $|\theta_i| \leq \alpha$, otherwise ${m}_i = 1$. Here $\alpha$ is a threshold value determined by the desired pruning percentage $p$, and $\mathbf{m_i}$ and $\theta{_i}$ are the $i^{\text{th}}$ elements of $\mathbf{m}$ and $\theta$, respectively.

    \item[4.] Repeat steps 2 and 3 for $j$ cycles or until the desired sparsity level is achieved.
    
\end{itemize}

The objective is to find a sparse subnetwork $\mathcal{F}(\mathbf{x}; \mathbf{m} \odot \theta)$ with accuracy $a' \geq a$, where $a$ is the accuracy of the original network $\mathcal{F}(\mathbf{x}; \theta)$, while pruning rounds $J$. 

\subsection{Challenges of LTH and Pruning Methods}
Proving the lottery ticket hypothesis for 3D shape classification models presents several challenges due to the geometric complexity and high dimensionality of 3D data:

\noindent\textbf{(1) Geometric Complexity:} 3D shapes exhibit intricate geometric structures and topological properties \cite{weinmann2017geometric, sheshappanavar2021dynamic, matveev2022def}. Let $\mathcal{G}$ be a set of geometric transformations (rotations, scaling), preserving the class label~\cite{zhou2019continuity, mo2024ric, kim2020rotation}. The 
subnetwork $\mathcal{F} (\mathbf{x}; \mathbf{m} \odot \theta)$ 
should satisfy:
$$
\mathcal{F}(\mathbf{g}(\mathbf{x}); \mathbf{m} \odot \theta) = \mathcal{F}(\mathbf{x}; \mathbf{m} \odot \theta), \quad \forall \mathbf{g} \in \mathcal{G}, \mathbf{x} \in \mathcal{X}
$$

\noindent\textbf{(2) High Dimensionality:} 
3D point cloud data presents unique challenges due to its high dimensionality and irregular structure \cite{tripathi2010high, georgiou2020survey}. Unlike 2D images, 3D point clouds can contain millions of points in irregular spatial arrangements, leading to larger input sizes and more complex feature representations. This often requires networks with more parameters $|\theta|$ to capture intricate spatial relationships and geometric features. The pruning objective for these larger networks is to minimize nonzero entries $\abs{\mathbf{m}}_1$ without compromising accuracy:
$$
\min_{\mathbf{m}} \abs{\mathbf{m}}_1 \quad \text{s.t.} \quad \mathcal{L}(\mathcal{F}(\mathbf{x}; \mathbf{m} \odot \theta)) \leq \mathcal{L}(\mathcal{F}(\mathbf{x}; \theta))
$$
where $\mathcal{L}$ is the loss function. This optimization is particularly challenging for 3D data due to the need to preserve complex spatial relationships and geometric features while significantly reducing the network size.

\noindent\textbf{(3) Structural Constraints:} 3D shapes can exhibit structural restrictions or relationships between different parts or components \cite{mitra2014structure,hillier2021three, bi2022deepismnet}. Let $\mathcal{C}$ be a set of structural constraints on the 3D shapes preserved by an overparametrized PCNN \cite{bronstein2017geometric,tekin2016structured}. If so, then the winning ticket subnetwork $\mathcal{F}(\mathbf{x}; \mathbf{m} \odot \theta)$ should preserve these constraints: 
$$
\mathcal{C}(\mathbf{x}) \Rightarrow \mathcal{C}(\mathcal{F}(\mathbf{x}; \mathbf{m} \odot \theta)), \quad \forall \mathbf{x} \in \mathcal{X}
$$
This formulation means that if a structural constraint $\mathcal{C}$ holds for an input 3D shape $\mathbf{x}$, then the same constraint should also hold for the output of the winning ticket subnetwork. For example, if an input shape has bilateral symmetry, the pruned network should still recognize and preserve this symmetry in its processing. It is important to maintain these constraints to preserve the semantic and geometric integrity of 3D shapes during classification tasks. We show a pruning method that selectively retains weights with the highest magnitudes, as these are more likely to encode important structural features.

\noindent\textbf{IMP and One-Shot Pruning:}
Let $\theta_l^{(j)}$ denote the weights of the $l^{th}$ layer in the PCNN $\mathcal{F}(\mathbf{x}; \theta)$ at pruning round $j$, and $m_l^{(j)}$ be the corresponding binary mask for that layer. The IMP process is as follows: Initialize the binary mask $\mathbf{m}^{(0)} = \mathbf{1}$ (all ones). For each round $j = 1, 2, \ldots, J$: Train the network $\mathcal{F}(\mathbf{x}; \mathbf{m}^{(j-1)} \odot \theta^{(j-1)})$ for one cycle. Determine the pruning threshold $\alpha_l^{(j)}$ for each layer $l$ based on the desired global or local pruning percentage $p$. Update the binary mask $m_l^{(j)}$ for each layer $l$ as follows: For global pruning \cite{yu2020easiedge}, we update the binary mask $m_l^{(j)}$ for layer $l$ at iteration $j$ in the following way. For each weight $\theta_l^{(j-1)}[i]$ at the index $i$, if its absolute value is less than or equal to the global threshold $\alpha^{(j)}$, we set the corresponding mask element $m_l^{(j)}[i]$ to 0 (pruned). Otherwise, we keep the mask element $m_l^{(j-1)}[i]$ at its previous value. Here, $\alpha^{(j)}$ is the global threshold determined by the $(1-p)^{th}$ percentile of the sorted magnitudes of all weights across all layers. For local pruning, similarly, $m_l^{(j)}[i] = 0$, if $|\theta_l^{(j-1)}[i]| \leq \alpha_l^{(j)} \ m_l^{(j-1)}[i],$ where $\alpha_l^{(j)}$ is the layer-wise threshold determined by the $(1-p)^{th}$ percentile of the sorted magnitudes of weights in layer $l$. We update the weights $\theta^{(j)} = \mathbf{m}^{(j)} \odot \theta^{(j-1)}$. The pruning terminates if the desired sparsity level is achieved, or $j = J$. The one-shot pruning \cite{chen2021only} is equivalent to IMP when $J$ = 1.

\noindent\textbf{Hypothesis:}
We hypothesize that the pruning methods (IMP, One-Shot) could address the challenges of geometric invariance, high dimensionality, sparsity, and structural constraint preservation in pruning by iteratively pruning the weights with the lowest magnitudes while retraining the network. By preserving the weights with the highest magnitudes, which are likely to capture salient geometric and structural features, the pruned network is expected to maintain geometric invariance $\mathcal{F}(\mathbf{g}(\mathbf{x}); \mathbf{m}^{(J)} \odot \theta^{(J)}) \approx \mathcal{F}(\mathbf{x}; \mathbf{m}^{(J)} \odot \theta^{(J)})$ and structural constraints $\mathcal{C}(\mathbf{x}) \Rightarrow \mathcal{C}(\mathcal{F}(\mathbf{x}; \mathbf{m}^{(J)} \odot \theta^{(J)}))$, while achieving high sparsity $\min_{\mathbf{m}^{(J)}} \sum_l |\mathbf{m}_l^{(J)}|_1$ and maintaining or improving accuracy. To validate the hypothesis, empirical evaluation is necessary, which is provided in the following sections.

\subsection{Transferability of 3D Subnetwork}

Transfer learning in 3D shape classification sometimes faces challenges due to limited data availability and domain specificity, unlike the 2D domain~\cite{hadgi2024supervise}. Moreover, applying LTH to a specific task often requires multiple pruning rounds to achieve the desired sparsity, making it computationally expensive to find the winning ticket for each task. On the other hand, in the 2D domain, LTH has demonstrated promising results in transferring tickets from models trained on large datasets to models trained on smaller datasets~\cite{morcos2019one}. Now, mathematically, let us consider two datasets, $\mathcal{D}_1$ and $\mathcal{D}_2$, both consisting of 3D shapes but with different distributions. Let $\mathcal{F}_1(\mathbf{x}; \theta_1)$ and $\mathcal{F}_2(\mathbf{x}; \theta_2)$ be the neural networks trained on $\mathcal{D}_1$ and $\mathcal{D}2$, respectively, for the task of 3D shape classification. The parameters $\theta_1$ and $\theta_2$ are initialized from the same distribution $\mathcal{D}_\theta$.
Suppose that we obtain a sparse subnetwork $\mathcal{F}_1(\mathbf{x}; \mathbf{m}_1 \odot \theta_1)$ from $\mathcal{F}_1(\mathbf{x}; \theta_1)$ using the iterative pruning and retraining approach, where $\mathbf{m}_1 \in \{{0, 1}\}^{|\theta_1|}$ is the binary mask that induces sparsity. The sparse subnetwork achieves an accuracy $a_1'$ on the validation set of $\mathcal{D}_1$, which is reasonably close to the accuracy $a_1$ of the original network $\mathcal{F}_1(\mathbf{x}; \theta_1)$.
To investigate the transferability of the sparse subnetwork, we initialize $\mathcal{F}_2(\mathbf{x}; \theta_2)$ with the parameters $\theta_2 = \mathbf{m}_1 \odot \theta_1$, where the non-zero elements of $\theta_2$ correspond to the winning tickets identified by the mask $\mathbf{m}_1$. We then fine-tune $\mathcal{F}_2(\mathbf{x}; \theta_2)$ on the dataset $\mathcal{D}_2$ using standard training procedures.
If the sparse subnetwork $\mathcal{F}_2(\mathbf{x}; \theta_2)$ can achieve an accuracy $a_2'$ on the validation set of $\mathcal{D}_2$ that is comparable to the accuracy $a_2$ of the original network $\mathcal{F}_2(\mathbf{x}; \theta_2)$, it would indicate that the winning tickets identified by the mask $\mathbf{m}_1$ on $\mathcal{D}_1$ are transferable to the task of classifying 3D shapes from the distribution $\mathcal{D}_2$.
Mathematically, the transferable ability can be quantified by the condition, $ a_2' \approx a_2, \quad \text{where} \quad \theta_2 = \mathbf{m}_1 \odot \theta_1$.
If this condition holds, it suggests that the sparse subnetwork obtained from $\mathcal{F}_1(\mathbf{x}; \theta_1)$ can be effectively transferred to the task of classifying 3D shapes from the distribution $\mathcal{D}_2$, thus reducing computational cost and training time while maintaining comparable performance.

\section{Experiments}

\noindent\textbf{Dataset:} We have investigated LTH on several synthetic and real-world 3D point cloud datasets. ModelNet40 \cite{wu20153d} is a synthetic Computer-aided design (CAD) generated dataset that contains 12,311 samples from 40 common objects. ScanObjectNN \cite{uy-scanobjectnn-iccv19} is a real-world dataset made by scanning real-world objects containing 15,000 samples of 3D shapes from 15 common categories. ShapeNetCore is a subset of the original ShapeNet \cite{chang2015shapenet} dataset where 51,300 instances are available from 55 categories of 3D shapes.

\noindent\textbf{Experimental Setups:} Unstructured point clouds are normalized and then augmented by random rotation and transformation as data processing steps for training the PCNN. After that, a fixed number of points, usually 1024 or 2048, are sampled from each point cloud shape data. Different approaches for pruning, such as IMP, global pruning, local pruning, and one-shot pruning, are used on each dataset. After successfully obtaining a reasonable sparsity without any accuracy drop, the PCNN is further pruned to an extreme sparsity level to investigate the true potential of the sparse network. 

During pruning, we investigated the possibility of making a PCNN up to 99\% sparse. The optimal performances are included according to the observation of the performance of the PCNN models vs. the pruning rate. We considered the optimal performance to be equal to or better than the unpruned model with possible sparsity or a negligible compromised accuracy with enormous sparsity. As part of our experiment, We pruned 10\%, 20\%, 30\%, and 40\% weights globally in an iterative manner, and by conducting an empirical testing and validation approach we found the threshold pruning rate for the one-shot global pruning. The number of train-prune-rewind cycles was different for each of our PCNN models. Using 100 cycles, we established the optimal performance and sparsity of the PointNet. While training, a batch size of 256 and the Adam optimizer with a learning rate of 0.0001 is used. We found that DGCNN requires 80 cycles to reach optimal performance, which is less than PointNet. We used the SGD optimizer with a learning rate of 0.1 and a small batch size of 32 for DGCNN. Applying the train-prune-rewind cycle on PointCNN showed a bit more complication than the other two models due to the structural difference. However, we managed to attain the optimal state within 80 cycles.  PointCNN is configured using a batch size of 128 and Adam optimizer, where the learning rate is set to 0.00001.

\noindent\textbf{Network:}
We have experimented with the pointwise  MLP-based method, PointNet \cite{qi2017pointnet}, point convolution-based method, PointCNN \cite{li2018pointcnn}, and graph-based methods, DGCNN\cite{wang2019dynamic}. These networks are established architectures to effectively learn from unstructured point cloud data inherent in 3D shape representations.

\noindent\textbf{Evaluation:} Following previous works \cite{girish2021lottery,chen2021lottery} on LTH, we evaluate our work based on testing accuracy and network pruning rate (sparsity) in percent.

\subsection{Main Results}

\begin{figure*}[t!]
\centering
\begin{center}
\includegraphics[width=1\linewidth]{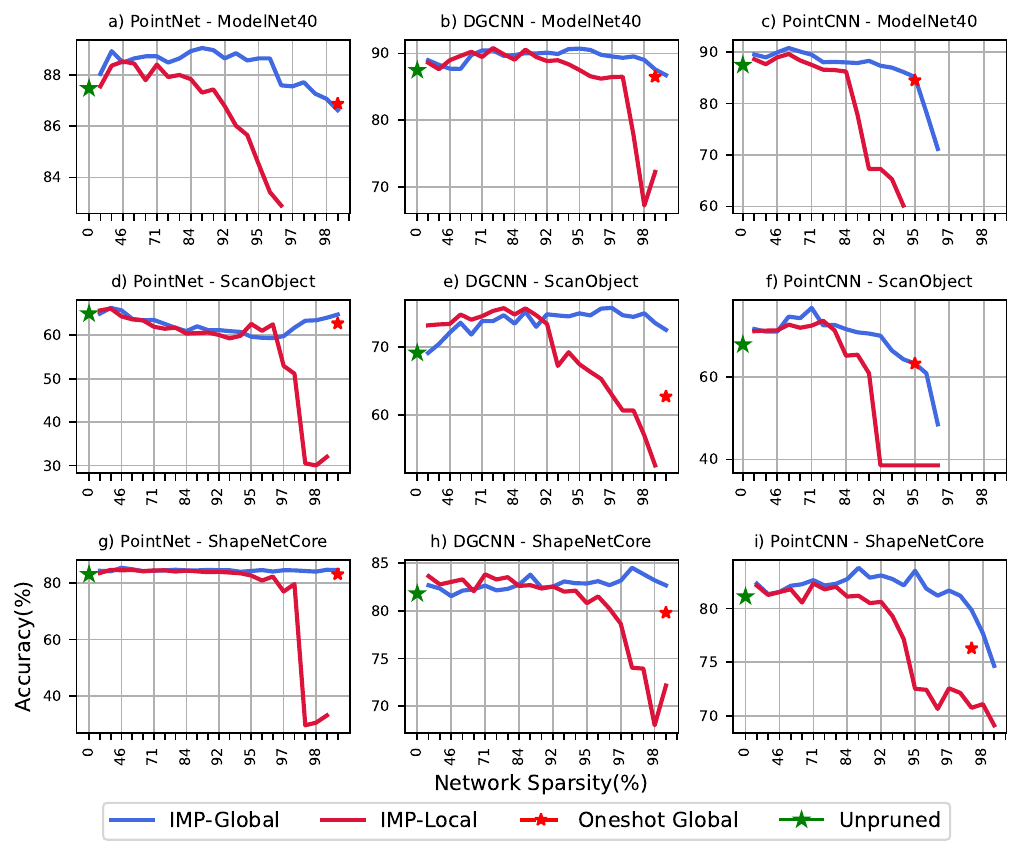}
\end{center}
\vspace{-2em}
\caption{\small The performance of lottery tickets discovered for 3D classification tasks across three model architectures: PointNet, DGCNN, and PointCNN, which handle point cloud data differently. Task-specific Iterative Magnitude Pruning (IMP) Global pruning outperforms the overparametrized baseline (0\% pruned) and IMP Local pruning by a wide margin. For efficiency, one-shot global pruning at 99\% sparsity exceeds all other methods evaluated. Each task's IMP method iteratively prunes 20\% weights to identify the winning tickets or subnetworks, achieving competitive or superior performance to the over parametrized models with reduced computation and parameters. }
\label{fig:overall_score}
\end{figure*}

\noindent\textbf{Sparse subnetworks:} We present our results of sparse subnetworks in Figure~\ref{fig:overall_score}. Our observations are the following:
\textbf{(1)} We notice the existence of highly sparse subnetworks with comparable or even superior accuracy to the original dense model. This finding is significant as the permutation-invariant nature of PCNN architecture makes it difficult to exploit spatial or temporal redundancies. \textbf{(2)} The sparse network achieved from global one-shot or IMP pruning maintains a high accuracy across various PCNN models and datasets. Results demonstrate the remarkably high pruning rates of up to 99\% for PointNet, PointCNN, and DGCNN architectures while preserving desirable accuracy levels. 
\textbf{(3)} Global pruning methods are more effective in identifying winning tickets at extreme sparsity levels than iterative pruning methods relying on local pruning criteria. Iterative local pruning methods struggle to uncover winning tickets at such a high pruning rate, suggesting that the weights (primarily responsible for the model's accuracy) are distributed globally rather than concentrated locally.
\textbf{(4)} Overparameterized PCNN models incorporate inherent redundancy, revealing that a small subset of weights can encode most of the learned knowledge.

\noindent\textbf{Transfer Learning Effectiveness:} Winning tickets can be generalized across related tasks or datasets \cite{morcos2019one} possibly because the identified subnetworks preserve common features to multiple tasks~\cite{meng2022contrastive}. We present our transfer results on 3D datasets in Table \ref{tab:transfer_results}. Our observations are:
\textbf{(1)} The winning tickets discovered through global one-shot pruning and IMP global pruning in PCNN models exhibit similar transferability across datasets. For example, the winning tickets obtained from models trained on the ShapeNetCore dataset can be successfully transferred to models trained on the ModelNet40 and ScanObjectNN datasets, achieving comparable performance to the original dense models trained on those datasets. 
\textbf{(3)} The transferability of winning tickets across datasets suggests that the identified subnetworks capture fundamental features relevant to the original and related tasks. This finding is particularly significant because it implies that a single winning ticket, once discovered, can be leveraged for efficient knowledge transfer and model adaptation across various point cloud datasets without retraining from scratch. 

\begin{table}[!t]
    \centering
    \begin{minipage}[t]{0.55\textwidth} 
        \centering
        \scalebox{0.55}{
            \begin{tabular}{ccccccccc} 
\toprule
\multirow{2}{*}{\begin{tabular}[c]{@{}c@{}}\textbf{Model}\\ \textbf{Accuracy(\%)}\end{tabular}} &             & \multicolumn{1}{l}{} & \multicolumn{3}{c}{\textbf{\textbf{ModelNet40}}} & \multicolumn{3}{c}{\textbf{ScanObjectNN}}         \\ 
\cline{4-9}
                                                                                                & \textbf{PR} & \textbf{Param (M)}   & \textbf{Base} & \textbf{IMP}  & \textbf{OneShot} & \textbf{Base} & \textbf{IMP}  & \textbf{OneShot}  \\ 
\midrule
PointNet                                                                                        & 0\%         & 3.400                & 87.5          & -             & -                & 68.2          & -             & -                 \\
Ours                                                                                             & 60\%        & 1.360                & -             & \textbf{88.2} & 87.8             & -             & \textbf{71.7} & 70.5              \\
Ours                                                                                             & 99\%        & \textbf{0.034}       & -             & 86.3          & 86.9             & -             & 64.7          & 62.7              \\ 
\midrule
DGCNN                                                                                           & 0\%         & 1.700                & 89.2          & -             & -                & 71.1          & -             & -                 \\
Ours                                                                                             & 60\%        & 0.680                & -             & 90.1          & \textbf{90.4}    & -             & \textbf{75.2} & 74.9              \\
Ours                                                                                             & 99\%        & \textbf{0.017}       & -             & 88.5          & 88.0             & -             & 64.6          & 60.6              \\ 
\midrule
PointCNN                                                                                        & 0\%         & 0.320                & 90.2          & -             & -                & 70.4          & -             & -                 \\
Ours                                                                                             & 60\%        & 0.128                & -             & \textbf{90.6} & 90.4             & -             & \textbf{73.2} & 72.9              \\
Ours                                                                                             & 99\%        & \textbf{0.003}       & -             & 87.4          & 86.9             & -             & 65.3          & 59.3              \\
\bottomrule
\end{tabular}
        }
        \vspace{-.75em} \caption{\small Sparse subnetworks, with sparsity levels of 99\% and 60\%, are transferred from a model trained on the larger ShapeNetCore dataset to models trained on the smaller ModelNet40 and ScanObjectNN datasets, respectively.}
        \label{tab:transfer_results}
    \end{minipage}
    \hfill
    \begin{minipage}[t]{0.42\textwidth} 
        \centering
        \scalebox{0.67}{
            \begin{tabular}{lcc}
                \toprule
                \textbf{Model} & \textbf{Param (M)}$\downarrow$ & \textbf{Accuracy(\%)}$\uparrow$\\
                \midrule
                PointCNN \cite{li2018pointcnn} & 0.320 & 92.2 \\
                Spherical CNNs \cite{esteves2018learning} & 0.500 & 88.9 \\
                Dense Point \cite{liu2019densepoint} & 0.530 & 93.2 \\
                KCNet \cite{shen2018mining} & 0.900 & 91.0 \\
                Point PN \cite{zhang2023parameter} & 0.800 & \textbf{93.8} \\
                \hline
                Ours (PointNet) & 0.034 & 87.1 \\
                Ours (DGCNN) & 0.017 & 86.3 \\
                Ours (PointCNN) & \textbf{0.012} & 85.4 \\
                \bottomrule
            \end{tabular}
        }
        \vspace{-0.6em} \caption{\small  Comparative analysis of the performance on the ModelNet40 dataset between existing less parametric models and highly sparse subnetworks extracted from over-parameterized models.}
        \label{tab:lessparametric}
    \end{minipage}
\end{table}

\noindent\textbf{Comparison with less parametric PCNN models:} Rather than reducing the parameters of a model by pruning, there exist  PCNN models, \cite{esteves2018learning}, \cite{liu2019densepoint}, \cite{shen2018mining}, \cite{zhang2023parameter} that use fewer parameters than the conventional larger parametric PCNN models. In Table \ref{tab:lessparametric}, we compare our pruned model's number of parameters and results with those models. The pruned version (with extremely few parameters) performs similarly to other less-parameter models.

\begin{figure}[!t]
    \centering
    \begin{minipage}{0.3\textwidth}
        \centering
        \includegraphics[width=\linewidth]{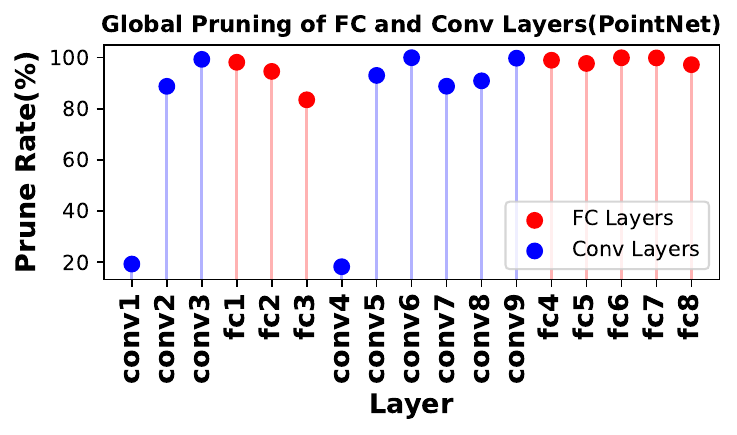}
    \end{minipage}
    \hfill
    \begin{minipage}{0.3\textwidth}
        \centering
        \includegraphics[width=\linewidth]{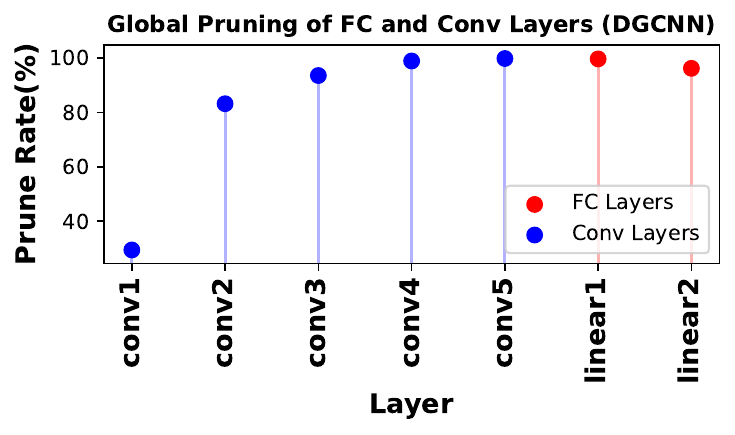}
    \end{minipage}
    \hfill
    \begin{minipage}{0.31\textwidth}
        \centering
        \scalebox{0.6}{\begin{tabular}{ccccc}
\hline
\multicolumn{5}{c}{\textbf{Ours (PointNet)}} \\ \hline
Conv & FC & Prune(\%) & Sparsity(\%) & Acc.(\%) \\
\hline
$\times$ & $\times$ & 0 & 0 & 87.5 \\
$\checkmark$ & $\times$ & 99 & 12 & 86.9 \\
$\times$ & $\checkmark$ & 99 & 87 & 88.3 \\
$\checkmark$ & $\checkmark$ & 99 & 99 & 86.8 \\
\hline
\end{tabular}}
    \end{minipage}
    \caption{\small The first two figures from the left illustrate the impact of global pruning on models such as PointNet and DGCNN. Based on the weight magnitude (importance), LTH prunes some Conv. layers with a significantly lower proportion of weights than FC layers. It tells LTH to automatically identify essential weights related to geometric transformations and other structural constraints of 3D data. The table on the right shows an ablation study for pruning the Conv. and FC layers separately.}
    \label{fig:ablation}
\end{figure}

\noindent\textbf{Ablation Studies:}
Our empirical analysis on PointNet and DGCNN for the ModelNet40 dataset, shown in Figure \ref{fig:ablation}, reveals that pruning a more significant proportion of the parameters of the FC layers leads to improved performance compared to the overparameterized network. Conversely, pruning only the conv layers results in a mere 12\% network sparsity at 99\% pruning, accompanied by a substantial degradation in performance that underscores the importance of preserving a representation from the Conv. layers compared to the FC layers. Also, upon analyzing the IMP method, it becomes evident that the convolutional layers preserve most of the weights.
 
\subsection{Discussion}
The demonstrated results highlight several notable observations and considerations concerning the architecture of PCNNs. First, it is evident that highly sparse subnetworks, or winning tickets, can be obtained within PCNN architectures, even at extreme pruning rates of up to 99\%. This finding contrasts with traditional models and datasets, where performance often starts to degrade at lower pruning rates, potentially due to the presence of rare features for specific classes in large datasets like ImageNet~\cite{corti2022studying,hossain2022colt}. 
Notably, further analysis reveals that even after aggressive pruning at 99\% rates, a significant portion of the weights in the convolutional layers of PCNN architectures remain intact. In contrast, many weights in the FC layers are pruned away. This finding suggests that the weights of the FC layers are less critical for overall performance than the Conv. layers responsible for extracting features from the point cloud data. The weights of the Conv layer are primarily essential to be preserved.
Moreover, the observation that one-shot global pruning at a 99\% pruning rate can achieve desirable accuracy. This implies that only 1\% of the highest magnitude weights are responsible for the model performance. These high-magnitude weights are predominantly concentrated in the convolutional layers, further emphasizing the significance of these layers in extracting essential features from point cloud data. These findings prioritize the preservation of convolutional layers while potentially eliminating or significantly compressing fully connected layers to develop more efficient architectures in the future.

\section{Conclusion}
We explore the lottery ticket hypothesis in PCNNs, revealing highly sparse subnetworks or "winning tickets" that maintain accuracy even at 99\% pruning rates (global pruning). Remarkably, these winning tickets display transferability across datasets, indicating that they capture fundamental point cloud features. The analysis underscores the importance of convolutional layers for feature extraction, while fully connected layers contribute significantly to model size without impacting performance. These findings lay the foundation for optimizing PCNN architectures, prioritizing convolutional layers, and achieving substantial model compression and efficient deployment, especially in resource-constrained settings. By leveraging transferable winning tickets and architectural insights, future research can drive applications such as autonomous systems and 3D computer vision advancements.

\bibliography{ref}
\end{document}